\documentclass[journal]{IEEEtran}

\makeatletter
\def\endthebibliography{%
  \def\@noitemerr{\@latex@warning{Empty `thebibliography' environment}}%
  \endlist
}

\usepackage{cite}
\usepackage{graphicx,psfrag}
\usepackage{bm}
\usepackage{epsfig}
\usepackage{amsmath}
\usepackage{array}
\usepackage{multirow}
\usepackage{color}
\usepackage{mathrsfs}
\usepackage{amsfonts}
\usepackage{caption}
\usepackage{subfigure}
\usepackage{amsthm}
\usepackage{amssymb}
\usepackage{algorithm}
\usepackage{algorithmic}
\usepackage{hyperref}
\usepackage{multirow}

\ifCLASSINFOpdf
\else
\fi
\usepackage{url}

\begin{document}
%
\title{Semantically Self-Aligned Network for Text-to-Image Part-aware Person Re-identification}
%
%
%

\author{Zefeng Ding,
        Changxing Ding,~\IEEEmembership{Member,~IEEE,}
        Zhiyin Shao,
        and Dacheng Tao,~\IEEEmembership{Fellow,~IEEE}
\thanks{Zefeng Ding and Zhiyin Shao are with the School of Electronic and Information Engineering, South China University of Technology,
381 Wushan Road, Tianhe District, Guangzhou 510000, China (e-mail: eezefengding@mail.scut.edu.cn; eezyshao@mail.scut.edu.cn).}
\thanks{Changxing Ding is with the School of Electronic and Information Engineering, South China University of Technology,
381 Wushan Road, Tianhe District, Guangzhou 510000, China. He is also with the Pazhou Lab, Guangzhou 510330, China (e-mail: chxding@scut.edu.cn).}
\thanks{Dacheng Tao is with the JD Explore Academy at JD.com, China (e-mail: taodacheng@jd.com).} }

\markboth{} 
{Shell \MakeLowercase{\textit{et al.}}: Bare Demo of IEEEtran.cls for IEEE Journals}
%



\maketitle

\begin{abstract}
Text-to-image person re-identification (ReID) aims to search for images containing a person of interest using textual descriptions. However, due to the significant modality gap and the large intra-class variance in textual descriptions, text-to-image ReID remains a challenging problem. Accordingly, in this paper, we propose a Semantically Self-Aligned Network (SSAN) to handle the above problems. First, we propose a novel method that automatically extracts part-level textual features for its corresponding visual regions. Second, we design a multi-view non-local network that captures the relationships between body parts, thereby establishing better correspondences between body parts and noun phrases. Third, we introduce a Compound Ranking (CR) loss that makes use of textual descriptions for other images of the same identity to provide extra supervision, thereby effectively reducing the intra-class variance in textual features. Finally, to expedite future research in text-to-image ReID, we build a new database named ICFG-PEDES. Extensive experiments demonstrate that SSAN outperforms state-of-the-art approaches by significant margins.
Both the new ICFG-PEDES database and the SSAN code will be released at \texttt{\url{https://github.com/zifyloo/SSAN}}.
\end{abstract}

\begin{IEEEkeywords}
Person re-identification, part-based models, image-text retrieval.
\end{IEEEkeywords}

%
\IEEEpeerreviewmaketitle

\section{Introduction}
\IEEEPARstart{T}{ext}-to-image person re-identification (ReID) refers to searching for images containing a person of interest (e.g. a missing child) based on natural language descriptions. It is a vital and powerful video surveillance tool when there are no probe images of the target person and only textual descriptions are available.
Compared with ReID works using pre-defined attributes~\cite{yang2020hierarchical, liu2018sequence, sarafianos2018deep},
textual descriptions contain significantly more information and therefore describe more diverse and more fine-grained visual patterns.
Unfortunately, the majority of the existing ReID literature focuses on image-based ReID, with text-to-image ReID still in its infancy \cite{wang2019beyond,leng2019survey}.

\begin{figure}[t]
\begin{center}
\includegraphics[width=1.0\linewidth]{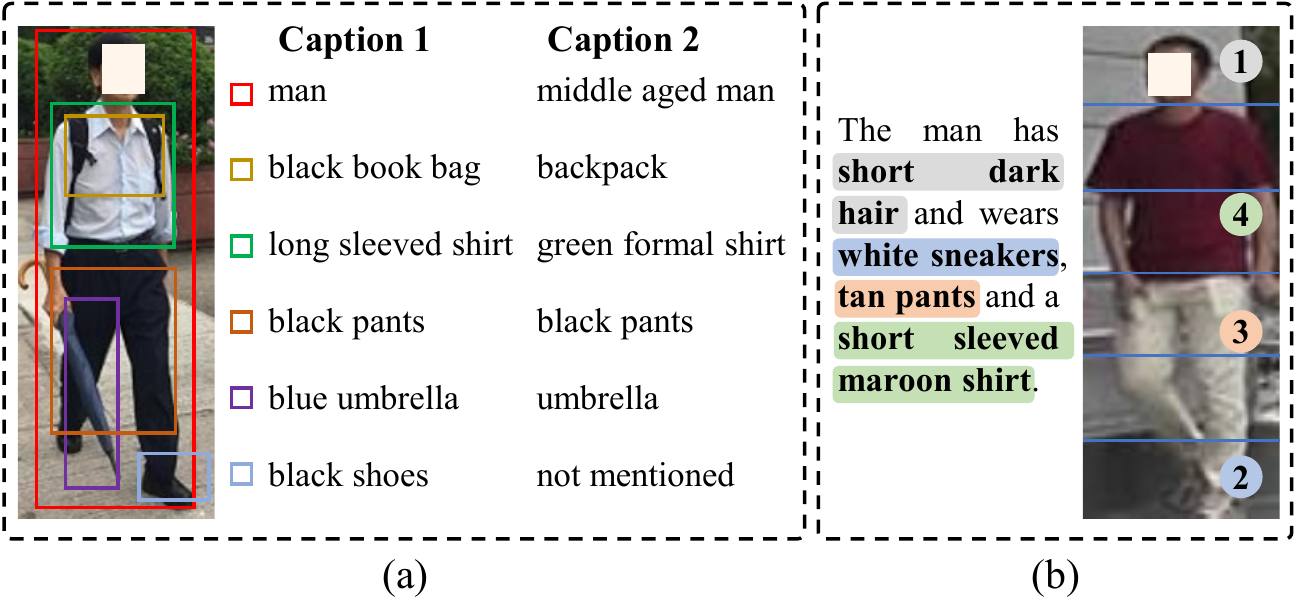}
\end{center}
   \caption{Textual descriptions are free-form, which introduces unique challenges to text-to-image ReID. (a) Textual descriptions for the same image may vary dramatically. (b) Body parts may be described in an arbitrary order. The numbers in this figure indicate the order in which the body parts are described.}
\label{fig:problems}
\end{figure}

Text-to-image ReID is much more challenging than image-based ReID. One of the main reasons for this is that textual descriptions are free-form, which creates two main problems. First, as illustrated in Fig.~\ref{fig:problems}(a), textual descriptions of the same image may vary dramatically, leading to large intra-class variance in textual features.
Second, body parts are usually well aligned after pedestrian detection; however, as illustrated in Fig.~\ref{fig:problems}(b), body parts may be described in an arbitrary order with various number of words, thereby introducing difficulty in extracting semantically aligned part-level features from both modalities.

Accordingly, cross-modal alignment is crucial for text-to-image ReID.
One popular cross-modal alignment strategy involves adopting attention models to acquire correspondences between body parts and words~\cite{li2017person,li2017identity,chen2018improving}.
However, this strategy depends on cross-modal operations for each image-text pair, which are computationally expensive~\cite{qu2020context}.
Another intuitive strategy involves splitting one textual description into several groups of noun phrases by using external tools, e.g. the Natural Language ToolKit \cite{loper2002nltk}.
Each group of noun phrases corresponds to one specific body part \cite{jing2020pose,niu2020improving,wang2020vitaa}.
The downside of this approach is that the quality of textual feature is sensitive to the reliability of the external tools.
Furthermore, the splitting operation destroys correlations between noun phrases, degrading the quality of textual feature.

In this paper, we propose a novel model, named Semantically Self-Aligned Network (SSAN), which efficiently extracts semantically aligned visual and textual part features.
SSAN does not split the textual description or perform cross-modal operations;
instead, it explores the relatively well aligned body parts in images as supervision and utilizes the contextual cues in language descriptions to achieve this goal.
More specifically, we first extract part-level visual features by splitting the feature maps from visual backbone into non-overlapping regions following \cite{sun2018beyond}.
Then, we process each textual description with a bidirectional long short-term memory network (Bi-LSTM) \cite{hochreiter1997long} to capture the relationships among the words.
With the aid of contextual cues, we infer the word-part correspondences using a Word Attention Module (WAM) based on the representation of each word.
Accordingly, we can obtain the raw part-level textural features with reference to the word-part correspondences.
The part-level visual and textual features are further refined by one shared $1\times1$ convolutional (Conv) layer between the two modalities.
Finally, by constraining the part-level features from both modalities to be similar, WAM is forced to make reasonable predictions and the semantic gap between the two modalities is reduced.

However, the above model ignores correlations between body parts. In textual descriptions, one noun phrase often covers several body parts (e.g. ``long dress''). Moreover, textual descriptions may specify certain spatial relationships between image regions (e.g. ``holding a bag''). Accordingly, we propose a multi-view non-local network (MV-NLN) based on the non-local attention mechanism \cite{wang2018non} to capture the relationships between body parts. Briefly, we first compute the similarity between the $k$-th part feature and each of the other part features in a shared embedding space via multi-view projection. Next, the similarity scores specify the strength of the interactions between the $k$-th part and the other parts. After interaction, the receptive field of each part feature extends to become more consistent with the noun phrases.

Furthermore, to overcome the intra-class variance in the descriptions, we propose a Compound Ranking (CR) loss. In traditional ranking loss \cite{faghri2017vse++}, positive pairs are composed of exactly matched image-text pairs. Motivated by the observation that one textual description can roughly annotate other images of the same identity, the CR loss adopts them to compose weakly supervised terms to optimize the network. However, the descriptive power of this rough annotation varies dramatically, depending on both the quality of the text and the difference in appearance between two images. Accordingly, we propose a strategy to adaptively adjust the margin for the new loss terms. The CR loss can be considered as a novel data augmentation method.

Finally, as there is only one large-scale database available (i.e., CUHK-PEDES \cite{li2017person}), we build a new database named Identity-Centric and Fine-Grained Person Description Dataset (ICFG-PEDES) for text-to-image ReID. Compared with CUHK-PEDES, our new database has three key advantages. First, its textual descriptions are identity-centric and fine-grained; in comparison, the textual descriptions contained in CUHK-PEDES are relatively short and may contain identity-irrelevant details (such as actions and backgrounds). More specifically, ICFG-PEDES has 58\% more words per caption on average than CUHK-PEDES. Second, the images included in ICFG-PEDES are more challenging, containing more appearance variability due to the presence of complex backgrounds and variable illumination \cite{wei2018person}. Third, the scale of ICFG-PEDES is larger: it contains 36\% more images than CUHK-PEDES.
We have made the ICFG-PEDES database publicly available, which is expected to further expedite research in text-to-image ReID.

We conduct extensive experiments on both the ICFG-PEDES and CUHK-PEDES databases. The results show that SSAN outperforms existing approaches by large margins. Moreover, SSAN has further advantages in terms of efficiency and ease of use.

The remainder of this paper is organized as follows. Related works are briefly reviewed in Section~\ref{related_work}. The SSAN model structure is introduced in Section~\ref{SSAN}.
Training and inference schemes are described in Section~\ref{Optimization}. Detailed experiments and their analysis are presented in Section~\ref{Experiments}. Finally, we conclude this paper in Section~\ref{Conclusion}.


\section{Related Work} \label{related_work}
In this section, we review the literature on (i) image-text retrieval, and (ii) part feature learning for image-based ReID.

\subsection{Image-text Retrieval}
Image-text retrieval is an important vision and language task. To obtain aligned features from the two modalities, early works typically projected holistic images and textual descriptions into a shared feature space \cite{frome2013devise,wang2016learning,gu2018look,klein2015associating,yan2015deep}. For example, Frome \emph{et al.} \cite{frome2013devise} proposed to learn linear transformations for image features and skip-gram word features, with a ranking loss employed to aid optimization. More recent approaches have further established accurate correspondences between image regions and words \cite{chen2020imram,huang2018learning,lee2018stacked,niu2017hierarchical,zhang2020context,wei2020multi,liu2020graph,wang2019camp}. For example, Zhang \emph{et al.} \cite{zhang2020context} considered both the interactions between the two modalities and semantic relationships in a single modality. Briefly, they adopted a region-to-word attention model to align cross-modal features, while also utilizing a second-order attention model to explore intra-modal correlations.

Due to its fine-grained nature, text-to-image ReID is one of the most challenging image-text retrieval tasks. Depending on the strategy used to align cross-modal features, existing works can be divided into two categories as follows:

\textbf{Novel Model Structures.} These works usually design various attention-based models to establish region-word \cite{li2017person,li2017identity,chen2018improving} or region-phrase \cite{jing2020pose,niu2020improving} correspondences. For example, as a region-word-based method, Chen \emph{et al.} \cite{chen2018improving} proposed a model that computed the affinity between each word and image region pair. They then selected the besting matching image region for each word. The final image-text affinity score was obtained via a word attention sub-network. Among the region-phrase-based methods, Niu \emph{et al.} \cite{niu2020improving} proposed a cross-modal attention model to align features from the two modalities at the global-to-global, global-to-local, and local-to-local levels in order to extract multi-granular features. However, these works require cross-modal operations for each image-text pair, which introduces a high computational cost \cite{qu2020context}. Recently, Wang \emph{et al.} \cite{wang2020vitaa} proposed an approach that is free from cross-modal operations. They aligned body parts with noun phrases with the help of an auxiliary segmentation layer and an external toolkit for sentence parsing. However, the textual features' quality relies on the reliability of the external tools.

\textbf{New Optimization Strategies.} These works utilize various objective functions to optimize text-to-image ReID models \cite{li2017identity,faghri2017vse++,sarafianos2019adversarial,zhang2018deep,zheng2020dual,liu2019deep,ge2019visual}. For example, Faghri \emph{et al.} \cite{faghri2017vse++} proposed a ranking loss to minimize the intra-class distance and maximize the inter-class distance. Sarafianos \emph{et al.} \cite{sarafianos2019adversarial} adopted an adversarial loss to drive textual features to be close enough to visual features such that a modality discriminator could not distinguish the two. Zheng \emph{et al.} \cite{zheng2020dual} proposed an instance loss that achieved better inter-modal alignment by sharing classifiers for the two modalities. However, these methods do not explicitly solve the significant intra-class variance problem in the textual modality.

\subsection{Part Feature Learning for Image-based ReID}
Part-based methods are popular for image-based ReID, as part-level representations include fine-grained features and thus effectively alleviate the overfitting risk associated with deep models~\cite{yao2019deep,sun2018beyond}.
Early methods extract part-level representations from uniformly partitioned feature maps produced by a CNN backbone~\cite{sun2018beyond}, as body parts were roughly aligned after pedestrian detection. In comparison, recent works promote the quality of part features by accounting for slight changes in part locations.

One intuitive strategy involves detecting body parts before feature extraction.
Methods in this category detect body parts by utilizing the outside tools \cite{kalayeh2018human,su2017pose,zhao2017spindle,xu2018attention,zheng2019pose,guo2019beyond} or the attention mechanism \cite{zhao2017deeply,li2018harmonious,liu2017hydraplus,wang2018mancs,tay2019aanet}.
For example, Guo \emph{et al.} \cite{guo2019beyond} detected body parts using a human parsing tool \cite{ruan2019devil} and important accessories with a self-attention scheme.
A few recent methods have attempted to bypass part detection during the testing stage~\cite{ding2020multi,zhang2019densely}; this has been implemented via more complicated training strategies, e.g. the teacher-student training scheme~\cite{zhang2019densely,ren2018factorized} or multi-task learning~\cite{ding2020multi}. For example, Zhang \emph{et al.} \cite{zhang2019densely} leveraged a 3D person model to construct a set of semantically aligned part images for each training image.
These part images enable a teacher model to learn semantically aligned part features.
Through the use of this teacher-student training scheme, the teacher model transfers the body part concepts to a student model,
which enables the student model to independently extract part-aware features in the testing phase.

While the methods discussed above are powerful, they also tend to utilize complex model architectures or training strategies to obtain high-quality part-level visual features.
In this paper, our research emphasizes the efficient extraction of high-quality part-level textual features.
Therefore, we adopt the simple uniform partitioning strategy for part-level visual feature extraction~\cite{sun2018beyond}.
It is worth noting that if the quality of part-level visual features is improved, we can accordingly obtain better part-level textual features by the proposed SSAN.

\begin{figure*}[!t] 
  \centering
  \includegraphics[width=1.0\linewidth]{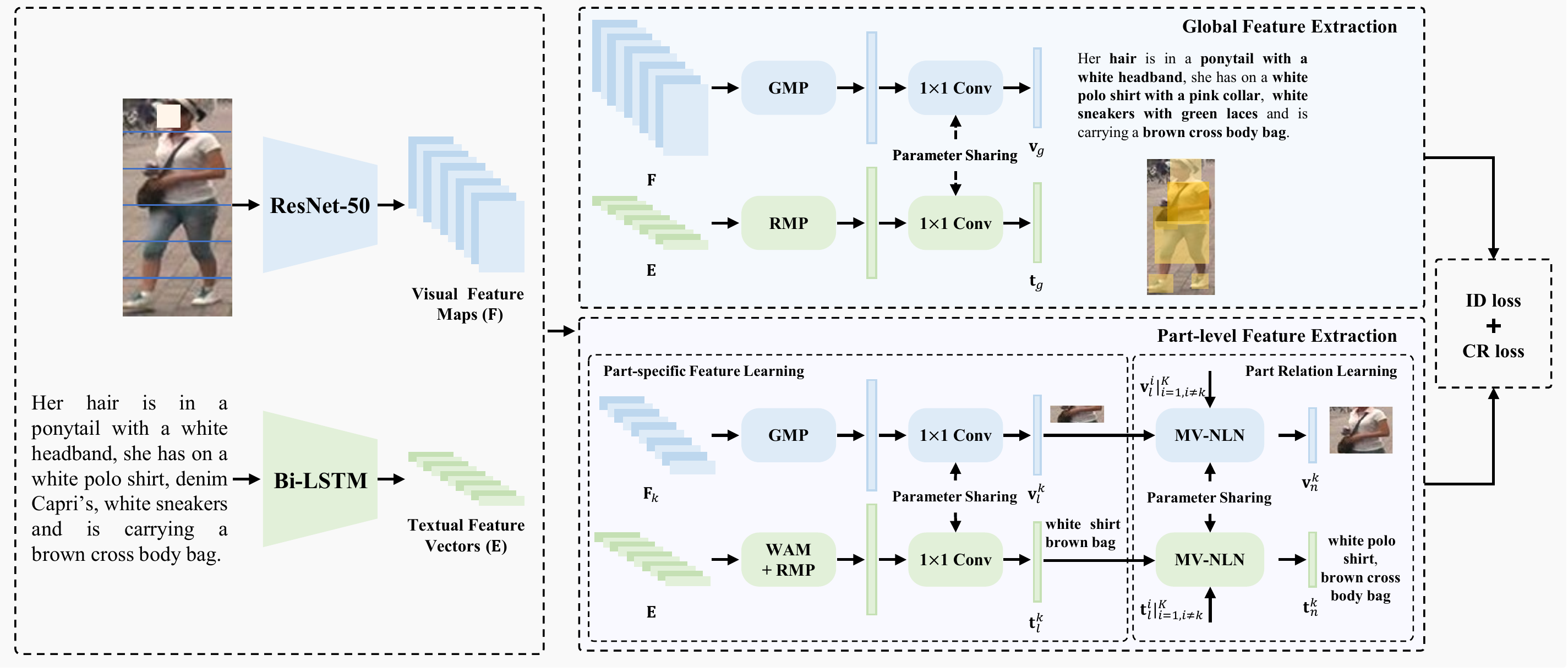}
  \caption{The SSAN model architecture. Based on the ResNet-50 and Bi-LSTM backbones, SSAN extracts global and part features respectively from both modalities. For simplicity, only the global branch and the $k$-th part branch are shown. $k$ is equal to 3 in this figure. The other $K$-1 part branches have the same structure as the $k$-th one. For the global branch, we adopt a weight sharing strategy on the last $1\times1$ Conv layer, which aligns the features from the two modalities more tightly in terms of semantics. Each part branch includes one Part-specific Feature Learning (PFL) module and one Part Relation Learning (PRL) module. The former module enables SSAN to automatically extract part-level features from both modalities, without using any external tools or cross-modal operations; the latter enables SSAN to capture the relationships between body parts so as to establish better semantic correspondences with noun phrases.}
\label{fig:structure}
\end{figure*}

\section{Semantically Self-Aligned Network} \label{SSAN}
The overall architecture of SSAN is illustrated in Fig.~\ref{fig:structure}. In what follows, we first introduce the backbone for representation extraction in Section~\ref{setion3.1} and then describe the branches for global and part feature learning in Sections~\ref{sec:globalFea} and~\ref{setion3.3}, respectively.

\subsection{Backbone}\label{setion3.1}

\noindent
\textbf{Visual Representation Extraction:} We adopt the popular ResNet-50 \cite{he2016deep} model as the visual feature extraction backbone. As shown in Fig.~\ref{fig:structure}, we first extract the feature maps ${\bf{F}}\in \mathbb{R}^{H \times W \times C}$, where $H$, $W$, and $C$ represent the height, width, and channel number of the feature maps, respectively. For the global branch, we directly utilize $\bf{F}$ to learn the global visual feature. For the part branches, we first uniformly partition ${\bf{F}}$ into $K$ non-overlapping parts ${\bf{F}}_{k} \in \mathbb{R}^{C\times H^{'} \times W}~(1\leq k \leq K)$, following \cite{sun2018beyond}. We then extract the part-level visual features from ${\bf{F}}_k$.

\noindent
\textbf{Textual Representation Extraction:} We build a matrix ${\bf{W}}_{e}\in \mathbb{R}^{V\times U}$ containing word embeddings of all unique words in the training set. Here, $V$ and $U$ denote the word embedding dimension and the number of unique words, respectively. Given a description $\bf{D}$ of length $n$, we obtain the word embedding ${\bf{x}}_{i} \in \mathbb{R}^{V}$  for its $i$-th word from ${\bf{W}}_{e}$.

To capture the relationships among the words, we adopt a bidirectional long short-term memory network (Bi-LSTM) \cite{hochreiter1997long} as our textual backbone. Bi-LSTM processes word embeddings from both ${\bf{x}}_1$ to ${\bf{x}}_n$ and ${\bf{x}}_n$ to ${\bf{x}}_1$ as follows:
\begin{equation}
\overrightarrow{\bf{h}}_i=\overrightarrow{LSTM}({\bf{x}}_i,\overrightarrow{{\bf{h}}}_{i-1}),
\end{equation}
\begin{equation}
\overleftarrow{\bf{h}}_i=\overleftarrow{LSTM}({\bf{x}}_i, \overleftarrow{{\bf{h}}}_{i+1}),
\end{equation}
where $\overrightarrow{\bf{h}}_i$, $\overleftarrow{\bf{h}}_i \in \mathbb{R}^{C}$ and represent the forward and backward hidden states of the $i$-th word, respectively. Next, the representation for the $i$-th word is defined as follows:
\begin{equation}
{\bf{e}}_{i} = \frac{\overrightarrow{\bf{h}}_{i}+\overleftarrow{\bf{h}}_{i}}{2}.
\end{equation}

Finally, we stack  ${\bf{e}}_{i}(1\leq i \leq n)$ to represent the textual description $\bf{D}$, as follows:
\begin{equation}
{\bf{E}} = [{\bf{e}}_{1}, {\bf{e}}_{2}, ..., {\bf{e}}_{n}],
\end{equation}
where ${\bf{E}} \in \mathbb{R}^{C\times  n}$.

\subsection{Global Feature Extraction}
\label{sec:globalFea}
Following the recent works \cite{jing2020pose,niu2020improving,wang2020vitaa}, SSAN aslo projects holistic visual and textual features into a common space. To obtain the global features, we first perform Global Max Pooling (GMP) on $\bf{F}$ and Row-wise Max Pooling (RMP) on $\bf{E}$. We then project the obtained features into a common feature space via a shared $1\times1$ Conv layer ${\bf{W}}_g$:
\begin{equation}
{\bf{v}}_{g} = {\bf{W}}_{g}GMP({\bf{F}}),
\end{equation}
\begin{equation}
{\bf{t}}_g = {\bf{W}}_{g}RMP({\bf{E}}),
\end{equation}
where ${\bf{W}}_{g} \in \mathbb{R}^{M \times C}$. ${\bf{v}}_g$, ${\bf{t}}_g \in \mathbb{R}^{M}$ and represent the global visual and textual features. Compared to previous methods \cite{jing2020pose,wang2020vitaa}, the weight-sharing strategy on ${\bf{W}}_g$ encourages $\bf{F}$ and $\bf{E}$ to be more tightly aligned in terms of semantics. In this way, our global branch outperforms that presented in \cite{jing2020pose,wang2020vitaa}, as demonstrated in experiment below.

Eventually, the similarity between global features of one image-text pair is denoted as follows:
\begin{equation}
S_{g}=\frac{{\bf{v}}_{g}^{T} {\bf{t}}_{g}}{\|{\bf{v}}_g\| \|{\bf{t}}_g\|}.
\label{equ:cosine_similarity}
\end{equation}

\subsection{Part-level Feature Extraction}\label{setion3.3}
Part-level representations are essential for ReID \cite{jing2020pose,niu2020improving,wang2020vitaa}. However, as explained in Fig.~\ref{fig:problems}, extracting part-level textual features is challenging. Therefore, we introduce the part branches in SSAN that efficiently extract semantically aligned part-level visual and textual features. Each part branch includes one Part-specific Feature Learning (PFL) module and one Part Relation Learning (PRL) module.

\subsubsection{Part-specific Feature Learning}

\begin{figure}[t]
\begin{center}
\includegraphics[width=1.0\linewidth]{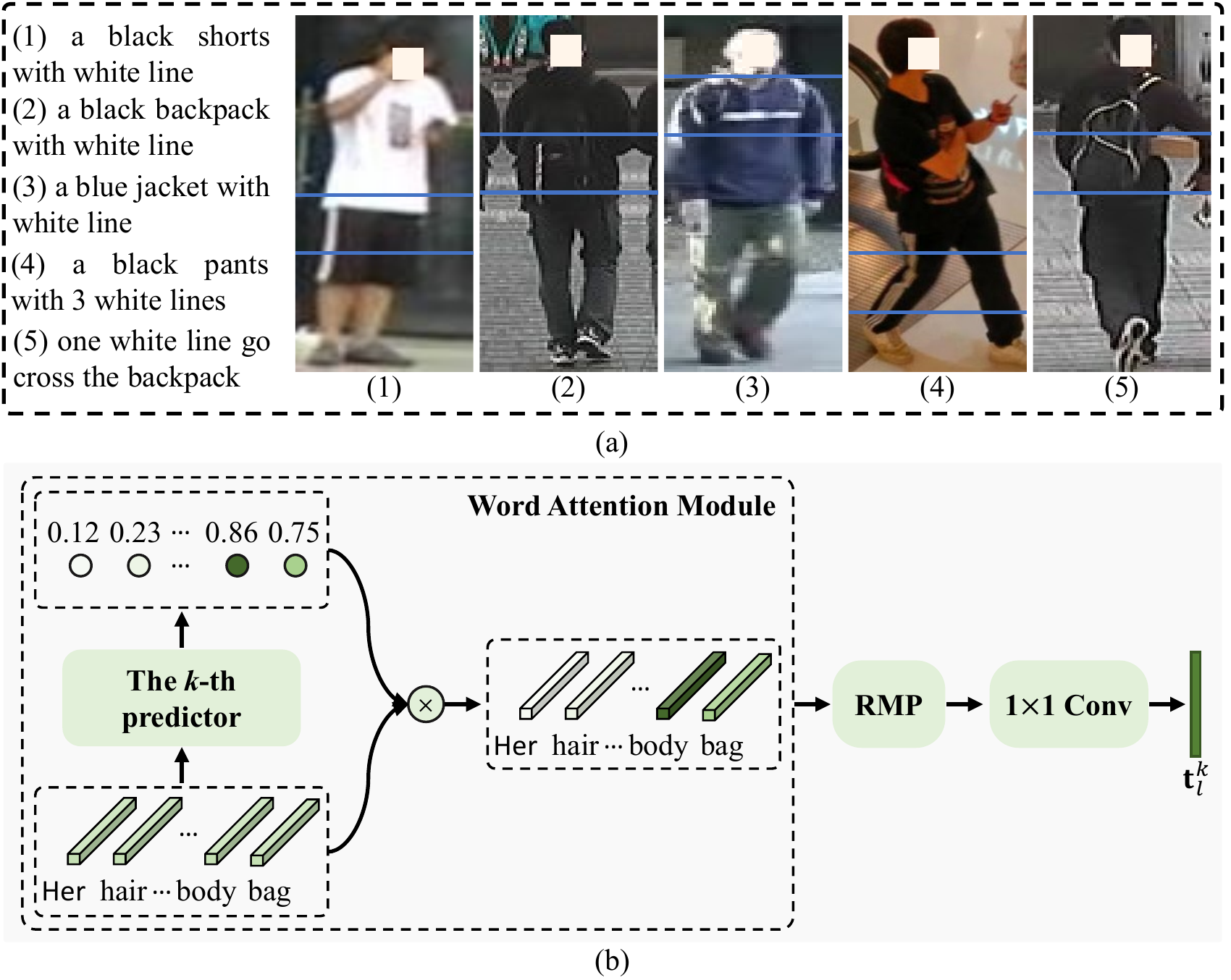}
\end{center}
   \caption{ (a) The same noun phrase (e.g., ``white line'') can be associated with different items of clothing or accessories. (b) The WAM model structure. We take the $k$-th part branch as an example. In this figure, $k$ is equal to 3.}
\label{fig:part_feature_learning}
\end{figure}

To obtain part-level textual features, existing approaches \cite{jing2020pose,niu2020improving,wang2020vitaa} usually first detect noun phrases using external tools  and then extract textual features for each part. However, this strategy breaks textual context. For example, as illustrated in Fig.~\ref{fig:part_feature_learning}(a), the noun phrase ``white line'' may relate to different items of clothing or accessories; without textual context, the correspondence between ``white line'' and body parts cannot be inferred.

It is therefore highly desirable to extract part-level textual features directly from the original textual description without external tools. As shown in Fig.~\ref{fig:part_feature_learning}(a), the body parts in images are usually well-aligned. Therefore, we utilize the well-aligned human body parts as supervision to facilitate achieving this goal. Furthermore, we argue that after the LSTM processing is complete, ${\bf{e}}_i$  will have obtained contextual cues that can be used to infer which part the $i$-th word corresponds to. Accordingly, we propose the following approach for the efficient extraction of semantically aligned part-level visual and textual features.

First, we introduce the Word Attention Module (WAM) to infer the word-part correspondences. As illustrated in Fig.~\ref{fig:part_feature_learning}(b), we predict the probability that the $i$-th word belongs to the $k$-th part as follows:
\begin{equation}
{{s}_i^k} = \sigma({\bf{W}}_{p}^k{{\bf{e}}_i}),
\end{equation}
where ${s}_i^k$ denotes the probability and $\sigma$ stands for the sigmoid function. ${\bf{W}}_{p}^{k} \in \mathbb{R}^{1 \times C}$. We modify  $\bf{E}$ to represent the textual description for the $k$-th part as follows:

\begin{equation}
{\bf{E}}_k = [{{s}_1^k}{\bf{e}}_1,{{s}_2^k}{\bf{e}}_2, ..., {{s}_n^k}{\bf{e}}_n].
\end{equation}

Second, as illustrated in Fig.~\ref{fig:structure}, we obtain the visual features for the $k$-th part by feeding ${\bf{F}}_k$ into one GMP layer and one $1\times 1$ Conv layer. Similarly, we generate the $k$-th part-level textual features by performing RMP on ${\bf{E}}_k$ and feed it to the same $1\times 1$ Conv layer as ${\bf{F}}_k$. Formally,

\begin{equation}
{{\bf{v}}_l^k} = {\bf{W}}_{l}^k GMP({{\bf{F}}_k}),
\end{equation}
\begin{equation}
{{\bf{t}}_l^k} = {\bf{W}}_l^k RMP({\bf{E}}_k),
\end{equation}
where ${\bf{W}}_{l}^{k} \in \mathbb{R}^{M \times C}$ and denotes the parameters of the shared $1\times 1$ Conv layer. ${\bf{v}}_l^k$, ${\bf{t}}_l^k \in \mathbb{R}^{M}$ and represent the visual and textual features for the $k$-th part, respectively.

\begin{figure}[t]
\begin{center}
\includegraphics[width=1.0\linewidth]{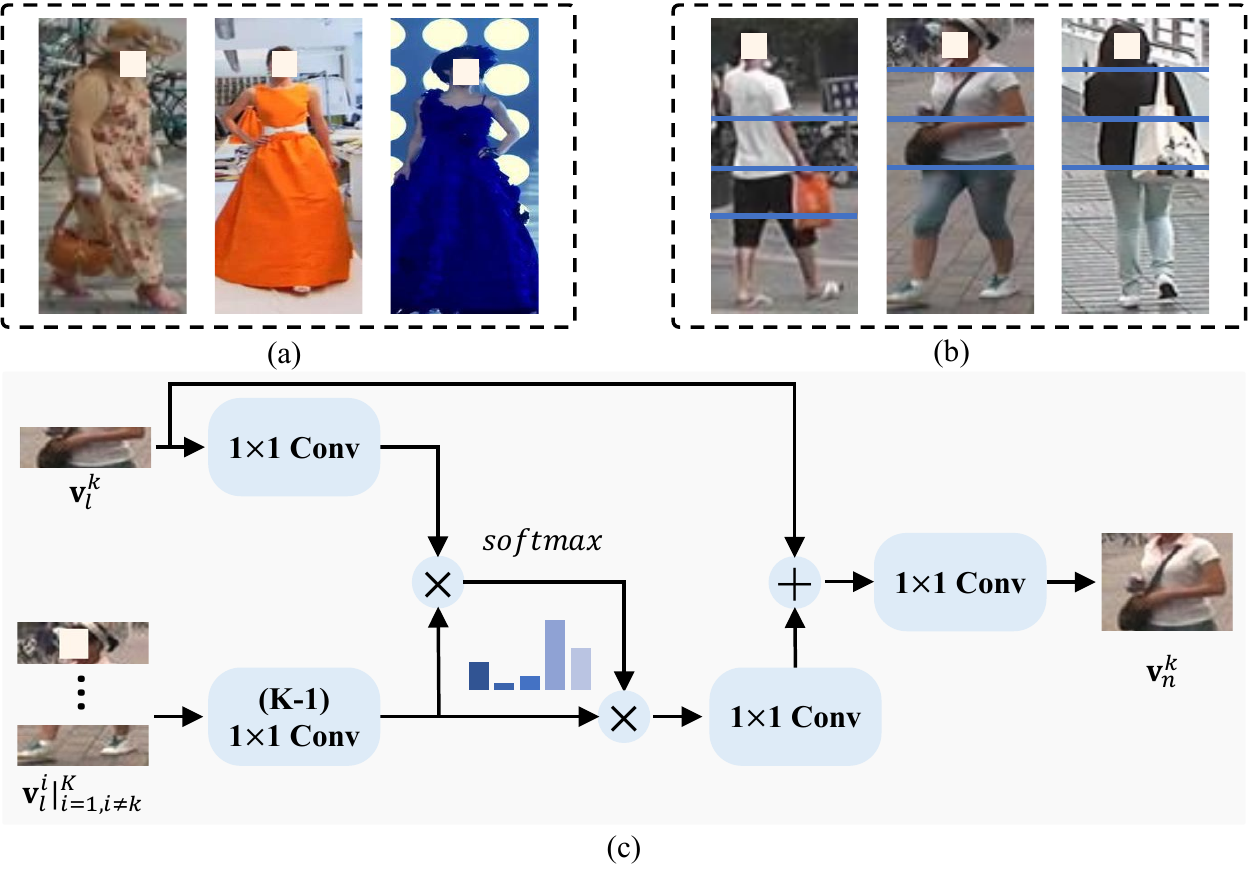}
\end{center}
   \caption{(a) One noun phrase may cover two or more equally divided body parts, e.g. ``long dress''. (b) Textural descriptions may include relationships between parts or image regions, e.g. ``holding a bag''. (c) The MV-NLN model structure. We take the $k$-th visual part feature ${\bf{v}}_{l}^{k}$ as an example. In this figure, $k$ is equal to 3.}
\label{fig:non-local}
\end{figure}

By constraining ${\bf{v}}_l^k$ and ${\bf{t}}_l^k$  to be both discriminative and similar, WAM is forced to make reasonable predictions about word-part correspondences. It is worth noting that one word may correspond to several parts, as explained in Fig.~\ref{fig:non-local}(a). Moreover, the shared $1\times1$ Conv layer is forced to select elements relevant to the $k$-th part from ${\bf{F}}_k$ and ${\bf{E}}_k$. In this way, we can obtain semantically aligned part-level visual and textual features without any external tools.

Finally, the similarity between the part-level features of one image-text pair is denoted as follows:
\begin{equation}
S_{l}=\frac{{\bf{v}}_l^T {\bf{t}}_l}{\|{\bf{v}}_l\| \|{\bf{t}}_l\|},
\label{equ:cosine_similarity}
\end{equation}
where ${\bf{v}}_l$ and ${{\bf{t}}_l} \in \mathbb{R}^{KM} $. They are obtained by concatenating the $K$ part-level visual and textual features, respectively.

\subsubsection{Part Relation Learning}
The equal partition strategy on $\bf{F}$ is effective for image-based ReID \cite{sun2018beyond,yao2019deep}. However, it maybe suboptimal for text-to-image ReID, as one phrase may cover two or more equally divided parts (e.g. ``long dress'' in Fig.~\ref{fig:non-local}(a)). Moreover, textual descriptions may specify relationships between parts (e.g. ``holding a bag'' and ``a bag cross chest'' in Fig.~\ref{fig:non-local}(b)). In this case, the correlation between parts is vital for differentiating the two phrases.

Here, we introduce the multi-view non-local network (MV-NLN) to address these problems. In the following, we take the $k$-th visual part feature as an example. As illustrated in Fig.~\ref{fig:non-local}(c), we first compute the similarity between ${\bf{v}}_l^k$ and ${\bf{v}}_l^i(i\neq k)$ in a shared embedding space via multi-view projections:
\begin{equation}
S_{ki}=\frac{{{\theta}_k({\bf{v}}_l^k)}^T {\phi}_i({\bf{v}}_l^i)}{\|{\theta}_k({\bf{v}}_l^k)\| \|{\phi}_i({\bf{v}}_l^i)\|},
\end{equation}
where ${\theta}_k({\bf{v}}_l^k) = {\bf{W}}_{\theta}^{k} {\bf{v}}_l^k$, ${\phi}_i({\bf{v}}_l^i) = {\bf{W}}_{\phi}^{i} {\bf{v}}_l^i$. ${\bf{W}}_{\theta}^{k}$, ${{\bf{W}}_{\phi}^{i}} \in \mathbb{R}^{M^{'} \times M}$. Then, the interaction strength between the $k$-th visual part feature and the other $K$-1 part features can thus be denoted as follows:

\begin{equation}
\alpha_{ki} = \frac{exp({S}_{ki})}{\sum_{i=1,i\neq k}^{K} exp({S}_{ki})},
\end{equation}
and $\alpha_{ki}$ is utilized to aggregate the $K$-1 part features:

\begin{equation}
{\bf{v}}_{l_{in}}^{k} = {\bf{W}}_{\gamma}^{k}(\sum_{i=1,i\neq k}^{K} {\alpha_{ki}{\phi}_i({\bf{v}}_l^i)}).
\end{equation}

Finally, the part-level visual features produced by MV-NLN can be denoted as follows:
\begin{equation}
{\bf{v}}_n^k = {\bf{W}}_{n}^k ({\bf{v}}_l^k + {\bf{v}}_{l_{in}}^{k}),
\end{equation}
where ${\bf{W}}_{n}^k \in \mathbb{R}^{N \times M}$ and ${\bf{W}}_{\gamma}^{k} \in \mathbb{R}^{M \times M^{'}}$.

Similar to the visual features, we also process the part-level textual features by using MV-NLN to capture their correlations. Note that the  parameters of MV-NLN are shared between the two modalities. Similar to $S_{g}$ and $S_{l}$, we adopt the cosine metric to evaluate the similarity between the features produced by MV-NLN for one image-text pair:
\begin{equation}
S_{n}=\frac{{\bf{v}}_n^T {\bf{t}}_n}{\|{\bf{v}}_n\| \|{\bf{t}}_n\|},
\label{equ:cosine_similarity}
\end{equation}
where ${\bf{v}}_n$ and ${\bf{t}}_n$ are obtained by concatenating the $K$ part-level visual and textual features produced by MV-NLN.

\section{Optimization} \label{Optimization}
The popular ranking loss \cite{faghri2017vse++} applies a constraint such that the intra-class similarity score must be larger than the inter-class similarity by a margin of $\alpha$ as follows:
\begin{equation}
\begin{aligned}
{L}_{r} & = max(\alpha- {S}({\bf{I}}_{p}, {\bf{D}}_{p}) + {S}({\bf{I}}_{p}, {\bf{D}}_{n}), 0)\\
& + max(\alpha-{S}({\bf{I}}_{p}, {\bf{D}}_{p}) + {S}({\bf{I}}_{n}, {\bf{D}}_{p}), 0),
\end{aligned}
\end{equation}
where ${\bf{I}}_p$ and ${\bf{D}}_p$ are drawn from a matching image-text pair. ${\bf{D}}_n$ and ${\bf{I}}_n$ denote the hardest negative text for ${\bf{I}}_p$ and the hardest negative image for ${\bf{D}}_p$  in a mini-batch, respectively. However, as shown in Fig.~\ref{fig:problems}, textual descriptions are flexible. The above ranking loss utilizes only the matching image-text pairs to compose positive pairs, which may result in a risk of overfitting.

\begin{figure}[t]
\begin{center}
\includegraphics[width=1.0\linewidth]{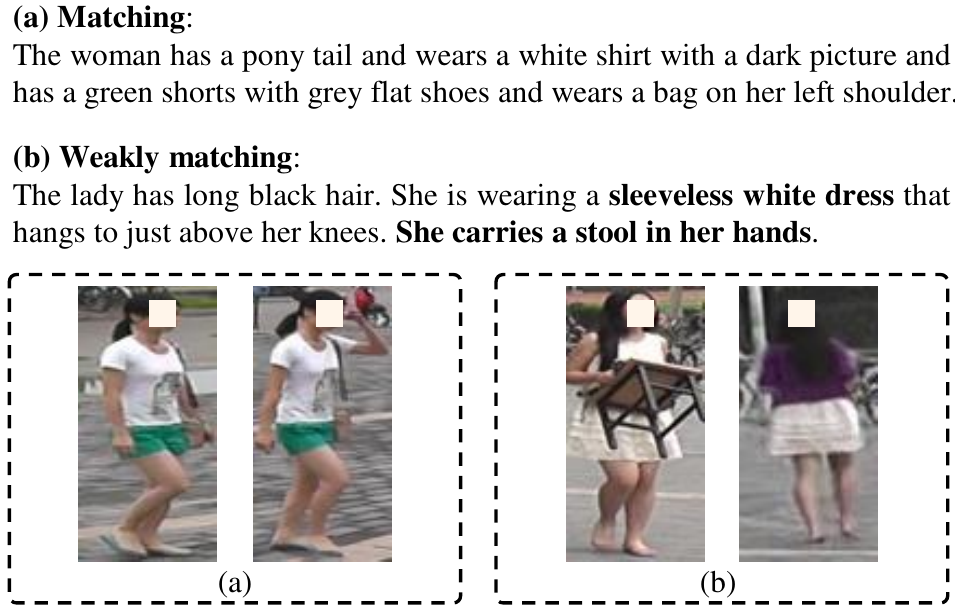}
\end{center}
   \caption{Textual descriptions can roughly annotate the other images of the same identity. However, their descriptive power varies dramatically.  (a) The textual description for the left image is also nearly perfect for the right one. (b) The textual description for the left image is only weakly relevant to the right image. }
\label{fig:hr_loss}
\end{figure}

As Fig.~\ref{fig:hr_loss} illustrates, textual descriptions can approximately annotate other images of the same identity; in other words, each textual description can be considered as a coarse caption for the other images of the same identity. Inspired by this observation, we propose a Compound Ranking (CR) loss that includes both strong and weak supervision terms. The positive pairs in strong supervision terms are drawn from image-text pairs that exactly match. By contrast, the positive pairs in weak supervision terms are composed of one image and the textual description for another image of the same identity. In this way, the CR loss exploits more diverse textual descriptions for each training image, which acts as a data augmentation strategy. Formally, the CR loss is defined as follows:
\begin{equation}
\begin{aligned}
{L}_{cr} &= max({\alpha}_1-S({\bf{I}}_{p}, {\bf{D}}_{p})+S({\bf{I}}_{p},{\bf{D}}_{n}), 0)\\
&+max({\alpha}_1- S({\bf{I}}_p, {\bf{D}}_p)+S({\bf{I}}_n,{\bf{D}}_{p}), 0)\\
&+\beta\cdot max({\alpha}_2-S({\bf{I}}_p,{\bf{D}}^{'}_p)+ S({\bf{I}}_p,{\bf{D}}_n), 0)\\
&+\beta\cdot max({\alpha}_2-S({\bf{I}}_p,{\bf{D}}^{'}_p)+ S({\bf{I}}_n,{\bf{D}}^{'}_{p}), 0),
\end{aligned}
\end{equation}
where ${\bf{D}}^{'}_p$ refers to the textual description for another image of the same identity as ${\bf{I}}_p$. ${\alpha}_1$ and ${\alpha}_2$ indicate the margins. $\beta$ denotes the weight for the weak supervision terms.

However, as illustrated in Fig.~\ref{fig:hr_loss}, the descriptive power of ${\bf{D}}^{'}_p$ for ${\bf{I}}_p$ varies due to the rich intra-class variance in image appearances, indicating that a fixed margin in the two weak supervision terms may be suboptimal. To overcome this problem, we propose the following strategy to adaptively adjust the value of ${\alpha}_2$:
\begin{equation}
{\alpha}_{2} = (\lambda + 1)  \frac{{\alpha}_{1}}{2},
\end{equation}
where
\begin{equation}
\lambda = min(\frac{S({\bf{I}}_p, {\bf{D}}^{'}_p)}{S({\bf{I}}_p,{\bf{D}}_p)}, 1).
\end{equation}

The CR loss and the popular ID loss \cite{zheng2020dual} are employed together to optimize the global features, the part features produced by PFL, and the part features produced by PRL, respectively. Note that the ID loss is imposed on each of the $K$ part features, while the CR loss is imposed on the concatenated $K$ part features. The weights of all loss terms for the three types of features are set to 1, 0.5, and 0.5.

In the testing stage, the overall similarity score between one image-text pair is the sum of $S_g$, $S_l$, and $S_n$.

\section{Experiments}\label{Experiments}
In this section, we conduct experiments on the CUHK-PEDES database \cite{li2017person}, as well as our newly constructed ICFG-PEDES database.
Following \cite{li2017person}, we adopt Rank-1, Rank-5, and Rank-10 accuracies to evaluate performance on both databases.

\begin{figure}[t]
\begin{center}
\includegraphics[width=1.0\linewidth]{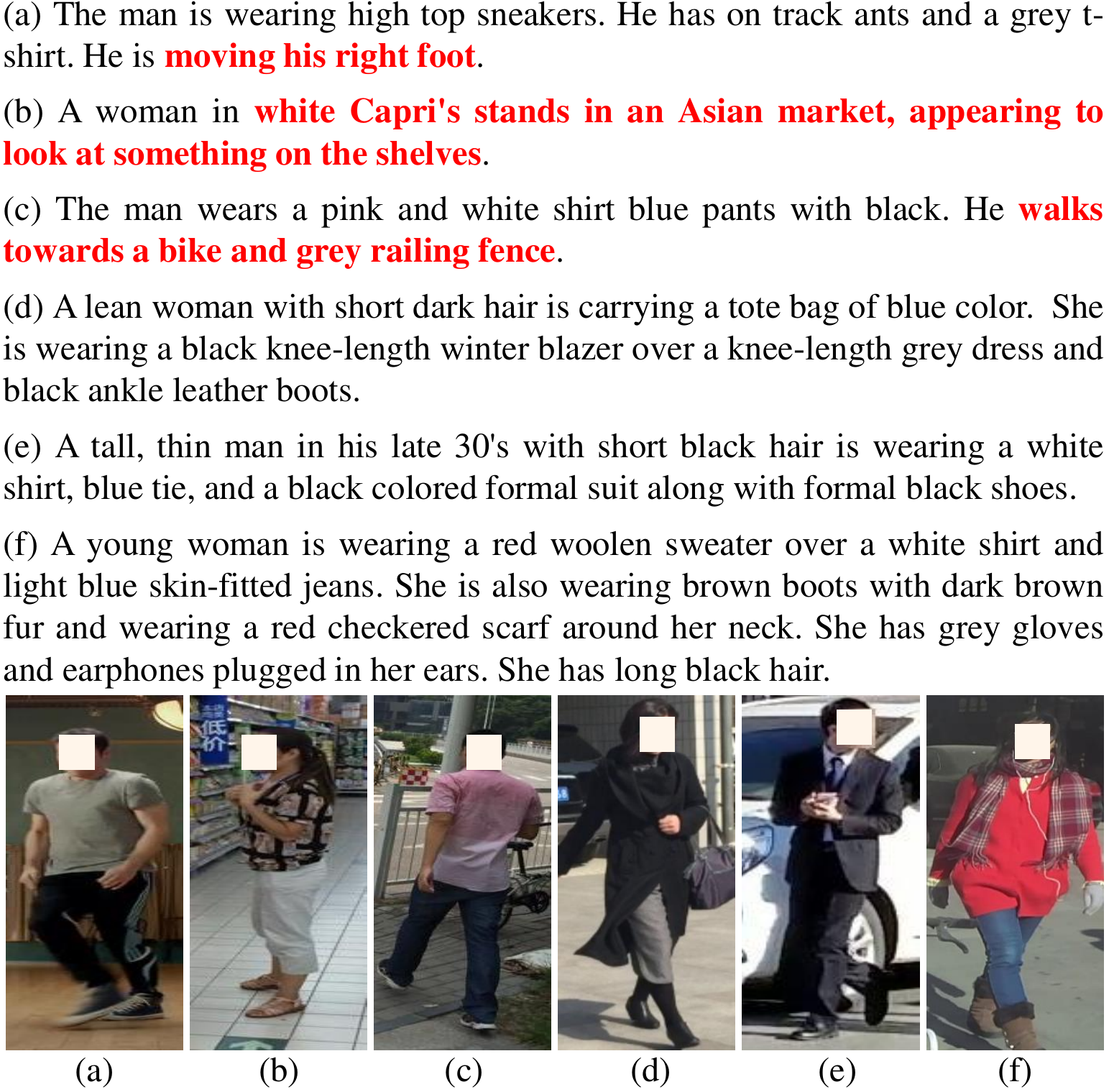}
\end{center}
   \caption{Comparisons between textual descriptions in the CUHK-PEDES and ICFG-PEDES databases. (a)-(c) Descriptions from CUHK-PEDES. The colored words are identity-irrelevant. (d)-(f) Descriptions from ICFG-PEDES show that the descriptions are more fine-grained in this dataset.}
\label{fig:cuhk-pedes}
\end{figure}

CUHK-PEDES contains 40,206 images and 80,412 textual descriptions for 13,003 identities with two captions per image. There are on average 23.5 words in each textual description. In line with the official evaluation protocol, the training set includes 34,054 images and 68,108 textual descriptions for 11,003 persons. The validation and test sets include data for 1,000 persons, respectively. The images and textual descriptions of the testing data make up the gallery set and probe set, respectively.

As there is only one large-scale database available for text-to-image ReID. It is hard for existing works to reliably verify the effectiveness of text-to-image ReID methods. Moreover, as illustrated in Fig.~\ref{fig:cuhk-pedes}, in CUHK-PEDES, textual descriptions may contain identity-irrelevant details (e.g. actions and backgrounds). Therefore, we construct a new database named ICFG-PEDES. The new database contains more identity-centric and fine-grained textual descriptions than CUHK-PEDES. It contains a total of 54,522 pedestrian images of 4,102 identities. All images are gathered from the MSMT17 database \cite{wei2018person}. There is one caption per image and there are on average 37.2 words for each description. The database contains a total of 5,554 unique words. Similar to the protocol in the original MSMT17 database, we divide ICFG-PEDES into a training set and a testing set: the former comprises 34,674 image-text pairs of 3,102 persons, while the latter contains 19,848 image-text pairs for the remaining 1,000 persons.

\begin{table*}[t]
\centering
\caption{Ablation Study on Each Component of SSAN}
\begin{tabular}{c|cccc|ccc|ccc}
\hline
  Dataset & \multicolumn{4}{c|}{Components} & \multicolumn{3}{c|}{CUHK-PEDES} & \multicolumn{3}{c}{ICFG-PEDES} \\
  \hline
  Metric        &Global & PFL & MV-NLN  & CR loss & Rank-1  &Rank-5 & Rank-10 & Rank-1 & Rank-5 & Rank-10 \\
  \hline
  \hline
  Baseline              &\checkmark   &-              &-              & -             & 54.68 & 75.42 & 82.73 & 49.02 & 69.05 & 76.55 \\
  + PFL                   &\checkmark   &\checkmark     &-              & -           & 59.26 & 78.56 & 85.32 & 52.58 & 71.70 & 78.75\\
  + PFL + PRL          &\checkmark   &\checkmark     &\checkmark     & -              & 60.59 & 79.37 & 86.01 & 53.53 & 72.38 & 79.23\\
  + PFL+1$\times 1$ Conv  &\checkmark   &\checkmark     &-              & -           & 59.36 & 79.06 & 85.56 & 52.63 & 72.05 & 78.90\\
  + CR loss               &\checkmark   &-              &-              &\checkmark   & 56.30 & 77.21 & 84.03 & 50.23 & 69.56 & 77.13\\
  \hline
  SSAN                   &\checkmark   &\checkmark     &\checkmark     &\checkmark    & \bfseries 61.37 & \bfseries 80.15 & \bfseries 86.73 & \bfseries 54.23 & \bfseries 72.63 & \bfseries 79.53 \\
  \hline
\end{tabular}
\label{tab:ablation}
\end{table*}


\subsection{Implementation Details}
Following a recent text-to-image ReID work \cite{wang2020vitaa}, we resize all images to $384\times128$ pixels and adopt random horizontal flipping for data augmentation.
To facilitate fair comparison with existing approaches, we adopt VGG-16 \cite{simonyan2014very} and ResNet-50 \cite{he2016deep} as the visual backbone, respectively;
both backbones are pre-trained on the ImageNet database \cite{russakovsky2015imagenet}.
For textual descriptions, we count the unique words to build a vocabulary in the training set.
Following previous methods \cite{wang2020vitaa}, $K$, $V$, $M$, $N$, and $\alpha_1$ are set to 6, 512, 1024, 512, and 0.2, respectively.
$M^{'}$ and $\beta$ are empirically set to 512 and 0.1, respectively.
During training, we adopt Adam \cite{kingma2014adam} as the optimizer. We set the batch size and number of epochs to 64 and 60, respectively. The learning rate is initially set as 0.001.

The baseline model in this section refers to the model in which we remove all part branches and the CR loss from SSAN; accordingly, the baseline model only extracts global features, as described in Section~\ref{sec:globalFea}. It adopts the cross-entropy loss and ordinary ranking loss~\cite{faghri2017vse++} for optimization.

\subsection{Ablation Study}
\label{sec:ablation}
In the following, we conduct ablation study to analyze the effectiveness of each key component of SSAN, i.e., Part Feature Learning module (PFL), Part Relation Learning module (PRL), and the CR loss. Experimental results on both databases are summarized in Table~\ref{tab:ablation}.

\subsubsection{Effectiveness of PFL}\label{ablation_pfl} In this experiment, we demonstrate the effectiveness of Part Feature Learning module.
Table~\ref{tab:ablation} presents the result of equipping the ``baseline''  with $K$ part branches adopting the PFL module alone.
As shown in Table~\ref{tab:ablation}, PFL promotes the Rank-1 accuracy of the baseline model by 4.58\% and 3.56\% on CUHK-PEDES and ICFG-PEDES, respectively.
The above comparisons demonstrate the effectiveness of PFL for automatically learning semantically aligned part-level features.

To further demonstrate the effectiveness of PFL in automatically extracting semantically aligned part-level textual features,
we compare its performance with two representative approaches, i.e., SCAN~\cite{lee2018stacked} and ViTAA~\cite{wang2020vitaa}, in Table~\ref{tab:pfl}.
In our experiments, SCAN adopts the same model architecture as SSAN to extract global features and part-level visual features.
It employs cross-modal operations to align body parts and words,
and subsequently extract semantically aligned part-level textual features.
ViTAA manually determines the correspondence between noun phrases and body parts.
We directly report the performance of ViTAA in its original paper~\cite{wang2020vitaa}.
Moreover, as our proposed WAM is able to predict the correspondence between body parts and words, we embed it into the framework of ViTAA to replace the manual annotations.
This model is denoted as ViTAA$^{*}$ in Table~\ref{tab:pfl}. From the comparisons between the above models, we can make the following observations.
First, PFL consistently outperforms both SCAN and ViTAA by considerable margins.
Second, with our WAM, ViTAA$^{*}$ significantly outperforms ViTAA significantly in terms of Rank-1 accuracy.
This means that WAM  more accurately predicts the correspondence between words and body parts.
One reason for this is that ViTAA~\cite{wang2020vitaa} only annotates the word-part correspondence for nouns.
In comparison, by utilizing the contextual cues, WAM is able to predict the correspondence for all words, e.g., nouns and adjectives.
The above comparisons further demonstrate the advantages of the proposed PFL.


\newcommand{\tabincell}[2]{\begin{tabular}{@{}#1@{}}#2\end{tabular}}
\newcommand{\thickhline}{
    \noalign {\ifnum 0=`}\fi \hrule height 1pt
    \futurelet \reserved@a \@xhline}

\begin{table}
\centering
\caption{Performance comparisons of different alignment methods on CUHK-PEDES. Rank-1 accuracy is adopted as the evaluation metric.}
\begin{tabular}{c|ccc}
    \hline
    Methods & Global+Part  & Part \\
    \hline
    \hline
    SCAN \cite{lee2018stacked}  & 55.86 & 54.24 \\
    ViTAA \cite{wang2020vitaa} & 55.97 & 39.26 \\
    ViTAA$^{*}$ & 57.58 & 56.73 \\
    PFL         & 59.26 & 58.02 \\
    \hline
\end{tabular}
\label{tab:pfl}
\end{table}


\subsubsection{Effectiveness of PRL} We next equip the baseline with both PFL and MV-NLN. As shown in Table~\ref{tab:ablation}, PRL further improves ReID performance by 1.33\% and 0.95\% in terms of Rank-1 accuracy on CUHK-PEDES and ICFG-PEDES respectively.

To ensure that SSAN benefits from the use of correlations between parts, we remove all layers except for the last $1 \times 1$ Conv of MV-NLN in PRL; the other experimental settings remain the same. As shown in Table~\ref{tab:ablation}, this model yields limited performance improvements on both databases, which indicate that the improvements achieved by PRL are due to its part-relation modeling ability.

\begin{figure}[t]
\begin{center}
\includegraphics[width=1.0\linewidth]{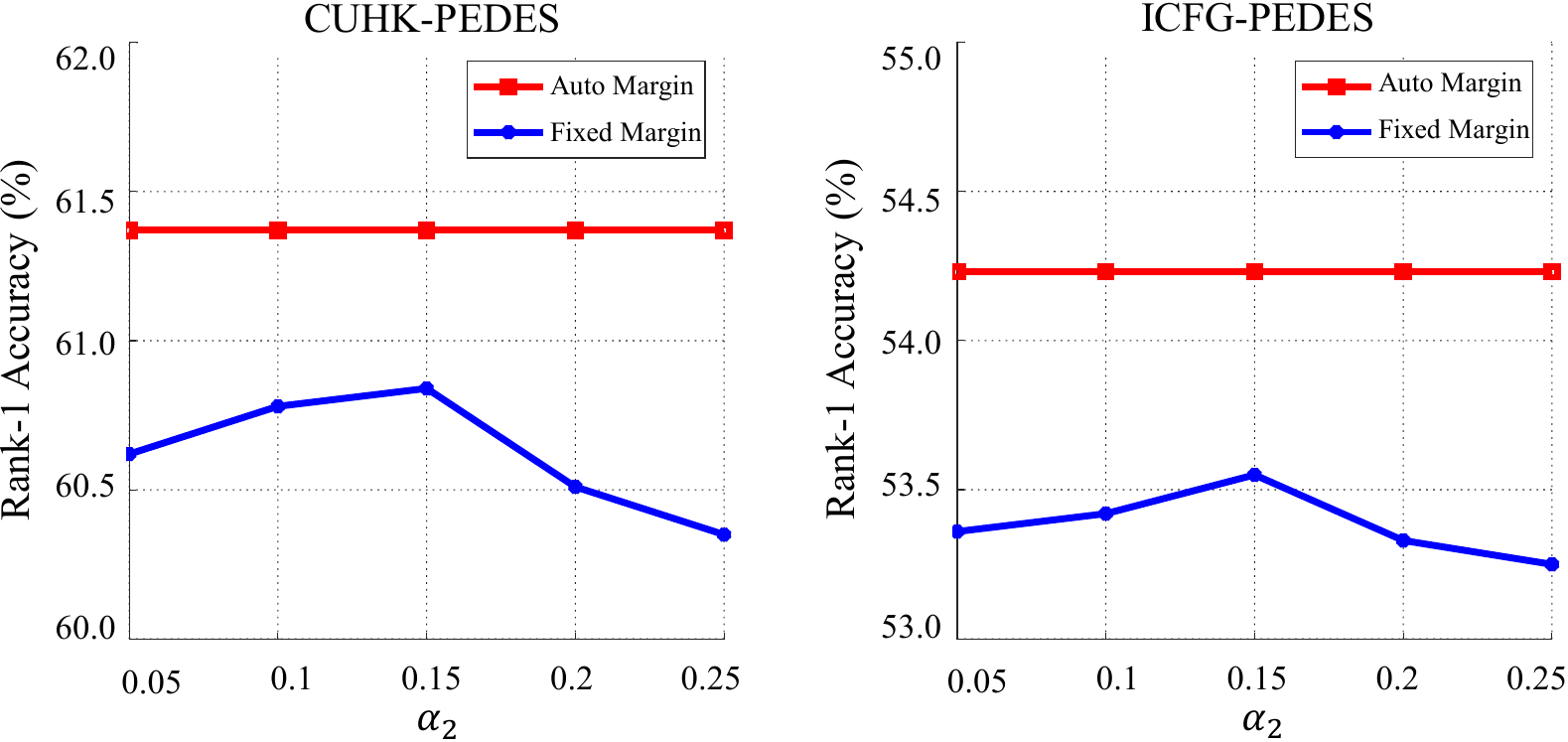}
\end{center}
   \caption{Comparisons between adaptive margins and fixed margins for the weak supervision terms in the CR loss.}
\label{fig:cr_margin}
\end{figure}

\subsubsection{Effectiveness of CR loss} We next explore the effectiveness of the CR loss. As shown in Table~\ref{tab:ablation}, the CR loss promotes the Rank-1 accuracy of the baseline model by 1.62\% and 1.21\% on CUHK-PEDES and ICFG-PEDES, respectively.

Furthermore, we equip the baseline model with PFL, MV-NLN, and the CR loss together, referred to as SSAN in Table~\ref{tab:ablation}. Compared with the model that uses PFL and MV-NLN alone, the CR loss promotes the Rank-1 accuracy by 0.78\% and 0.7\% on the two databases, respectively.

Finally, we provide comparisons between adaptive margins and fixed margins for the weak supervision terms in the CR loss.
Experimental results are illustrated in Fig.~\ref{fig:cr_margin}.
It is shown that SSAN with adaptive margins consistently outperforms SSAN with fixed margins on both databases.
Compared with fixed margins, adaptive margins are more flexible to the dramatic variation in the descriptive power of the rough annotations.
The above comparisons demonstrate the effectiveness of the proposed CR loss.

\subsection{Comparisons with State-of-the-Art Methods}

\subsubsection{Performance Comparisons on CUHK-PEDES}
We compare the performance of SSAN with state-of-the-art methods on the CUHK-PEDES database in Table~\ref{tab:cuhkCompare}.
Most approaches in this table are also part-based \cite{chen2018improving,jing2020pose,niu2020improving,wang2020vitaa}.
Depending on which backbone is adopted, we divide these approaches into two groups: VGG-16-based methods and ResNet-50-based methods.
We can make the following observations from Table~\ref{tab:cuhkCompare}.

First, thanks to the weight-sharing strategy, our baseline model performs better than that in ViTAA \cite{wang2020vitaa}. Without the weight-sharing strategy, our baseline achieves 53.12\% Rank-1 accuracy with the ResNet-50 backbone, which is approximately the same as that in ViTAA.

Second, SSAN significantly outperforms all existing approaches in terms of all metrics. More specifically, SSAN outperforms the most recent method ViTAA \cite{wang2020vitaa} by 5.4\% in terms of Rank-1 accuracy. It is worth noting that both ViTAA and SSAN adopt the same input image size and backbones. Of the VGG-16-based methods, SSAN outperforms TDE \cite{niu2020textual} by 3.15\% in terms of Rank-1 accuracy. Another advantage of SSAN is its ability to automatically extract semantically aligned part features from both modalities. By comparison, the other methods typically rely on external tools to extract part-level textual features~\cite{jing2020pose,niu2020improving,wang2020vitaa}. These comparisons demonstrate the effectiveness of SSAN.

\begin{table}
\centering
\caption{Performance Comparisons on CUHK-PEDES}
\begin{tabular}{c|c|ccc}
    \hline
      \multicolumn{2}{c|}{Methods} &Rank-1 & Rank-5 & Rank-10 \\
    \hline
    \hline
      \multirow{7}*{\rotatebox{90}{VGG-16}}
      & GNA-RNN \cite{li2017person} & 19.05 & - & 53.64 \\
      & IATVM \cite{li2017identity} & 25.94 & - & 60.48\\
      & PWM-ATH \cite{chen2018improving} & 27.14 & 49.45 & 61.02\\
      & PMA \cite{jing2020pose} & 47.02 & 68.54 & 78.06\\
      & MIA \cite{niu2020improving} &48.00 & 70.70 & 79.30\\
      & TDE \cite{niu2020textual}  & 52.37  & 75.26  & 83.31\\
      \cline{2-5}
      & \bfseries SSAN &{\bfseries 55.52} &{\bfseries 76.17} &{\bfseries 83.45} \\

    \hline\hline
      \multirow{10}*{\rotatebox{90}{ResNet-50}}
      & Dual Path \cite{zheng2020dual}   & 44.40 & 66.26 & 75.07\\
      & CMPM + CMPC \cite{zhang2018deep} & 49.37 & -     & 79.27\\
      & MIA \cite{niu2020improving}      & 53.10 & 75.00 & 82.90\\
      & PMA \cite{jing2020pose}          & 53.81 & 73.54 & 81.23\\
      & TDE \cite{niu2020textual}        & 55.25 & 77.46 & 84.56\\
      & SCAN \cite{lee2018stacked}       & 55.86 & 75.97 & 83.69\\
      & ViTAA \cite{wang2020vitaa}       & 55.97 & 75.84 & 83.52\\
      \cline{2-5}
      & Baseline in \cite{wang2020vitaa}  & 52.27 & 73.33 & 81.61\\
      & Baseline  in SSAN  & 54.68 & 75.42 & 82.73\\
      & \bfseries SSAN &{\bfseries 61.37} &{\bfseries 80.15} &{\bfseries 86.73} \\
    \hline
\end{tabular}
\label{tab:cuhkCompare}
\end{table}

\begin{table}
\centering
\caption{Performance Comparisons on ICFG-PEDES}
\begin{tabular}{c|c|ccc}
    \hline
      \multicolumn{2}{c|}{Methods} &Rank-1 & Rank-5 & Rank-10\\
    \hline
    \hline
      \multirow{6}*{\rotatebox{90}{ResNet-50}}
      & Dual Path \cite{zheng2020dual}   & 38.99 & 59.44 & 68.41\\
      & CMPM + CMPC \cite{zhang2018deep} & 43.51 & 65.44 & 74.26\\
      & MIA \cite{niu2020improving}      & 46.49 & 67.14 & 75.18\\
      & SCAN \cite{lee2018stacked}       & 50.05 & 69.65 & 77.21\\
      & ViTAA$^*$ \cite{wang2020vitaa}   & 50.98 & 68.79 & 75.78\\
      \cline{2-5}
      & \bfseries SSAN &\bfseries 54.23 & \bfseries 72.63 & \bfseries 79.53\\
    \hline
\end{tabular}
\label{tab:icfgCompare}
\end{table}

\subsubsection{Performance Comparisons on ICFG-PEDES}
We evaluate the performance of existing methods~\cite{zheng2020dual,zhang2018deep,niu2020improving,lee2018stacked} with open-source code on the new ICFG-PEDES database.
Furthermore, as shown in Table~\ref{tab:pfl}, ViTAA$^{*}$ both outperforms ViTAA and enjoys the advantage of independence on manual annotations.
We therefore report the performance of ViTAA$^{*}$ on ICFG-PEDES. Comparison results on the ICFG-PEDES database are summarized in Table~\ref{tab:icfgCompare}.
All methods in this table adopt the same backbone model, i.e., ResNet-50.
It is shown that SSAN outperforms all other methods by significant margins. For example, SSAN beats the SCAN model \cite{lee2018stacked} by 4.18\% in terms of Rank-1 accuracy. Moreover, when compared with ViTAA$^{*}$, SSAN still achieves a large performance gain by as much as 3.25\% in terms of Rank-1 accuracy.
These experimental results demonstrate the effectiveness of SSAN for text-to-image ReID.

\begin{table*}[t]
\centering
\caption{Performance Comparisons on CUHK-PEDES in the Cross-domain Settings}
\begin{tabular}{c|c|cc|cc|cc|cc}
    \hline
\multicolumn{2}{c|}{Dataset} & \multicolumn{2}{c|}{S$\rightarrow$C03 } & \multicolumn{2}{c|}{ S$\rightarrow$M} & \multicolumn{2}{c|}{S$\rightarrow$V} & \multicolumn{2}{c}{S$\rightarrow$C01}  \\
    \hline
\multicolumn{2}{c|}{Metric} & Rank-1 & Rank-5    & Rank-1 & Rank-5     & Rank-1 & Rank-5    & Rank-1 & Rank-5 \\

    \hline
    \hline
    \multirow{4}*{\tabincell{c}{SO}}

              & CMPM+CMPC \cite{zhang2018deep}   &42.3 &69.2   &63.4 &85.1   &57.8 &84.7   &44.8 &70.9  \\
              & MIA \cite{niu2020improving}      &49.0 &76.7   &66.2 &86.2   &55.1 &84.7   &50.2 &75.9   \\
              & SCAN \cite{lee2018stacked}       &50.2 &75.9   &64.2 &86.2   &55.1 &81.1   &48.2 &76.8   \\

              & \bfseries SSAN       &\bfseries 54.5 &\bfseries 78.5  &\bfseries 71.1 &\bfseries 88.6   &\bfseries 66.3 &\bfseries 89.3   &\bfseries 60.5 &\bfseries 81.3 \\

    \hline
    \hline
    \multirow{4}*{\tabincell{c}{ST}}
              & SPGAN \cite{deng2018image}        &44.7 &72.5   &63.3 &85.3   &60.7 &85.7   &45.3 &71.2 \\
              & ADDA  \cite{tzeng2017adversarial} &45.1 &72.8   &63.9 &85.7   &61.4 &86.0   &45.7 &71.6 \\
              & ECN  \cite{zhong2019invariance}   &45.8 &73.2   &64.3 &86.1   &62.5 &86.4   &46.6 &72.1 \\
              & MAN   \cite{jing2020cross}        &48.5 &74.8   &65.1 &87.4   &64.2 &87.2   &48.2 &73.2 \\
    \hline

\end{tabular}
\label{tab:CUHK-CD}
\end{table*}

\subsubsection{Performance Comparisons on CUHK-PEDES with Cross-domain Settings}
To further validate the generalization ability of SSAN, we carry out experiments on CUHK-PEDES with the cross-domain settings proposed in MAN~\cite{jing2020cross}.
The CUHK-PEDES database is divided into five independent subsets as five domains according to the source of the images,
i.e., CUHK03~\cite{li2014deepreid}, Market-1501 \cite{zheng2015person}, SSM \cite{xiao2016end}, VIPER \cite{gray2007evaluating}, and CUHK01 \cite{li2012human}.
SSM (S) is chosen as the source domain and there are four transfer tasks, namely, S$\rightarrow$C03 (CUHK03), S$\rightarrow$M (Market-1501), S$\rightarrow$V (VIPER), and S$\rightarrow$C01 (CUHK01). For each task, there are two different settings: (i) Source Only (SO). In this setting, one model is trained with only labeled source data.
(ii) Source and Target (ST). In this setting, one model is trained with the labeled source data and unlabeled target data.
Comparison results in the two settings are summarized in Table~\ref{tab:CUHK-CD}.
SCAN is implemented according to the description in Section~\ref{sec:ablation}. MIA is implemented based on its open source code. They extract both global and part features; therefore, the comparisons among SCAN, MIA, and SSAN are fair.
For all other methods in Table~\ref{tab:CUHK-CD}, we directly cite their performance reported in~\cite{jing2020cross}.
From these results, we can make the following observations.

First, in the SO setting, SSAN outperforms all other methods in Table~\ref{tab:CUHK-CD} by considerable margins for all four transfer tasks.
More specifically, SSAN outperforms SCAN \cite{lee2018stacked} by 4.3\%, 6.9\%, 11.2\%, and 12.3\% in terms of Rank-1 accuracy on the four transfer tasks, respectively.
This is because SCAN extracts part-level textual features by performing cross-modal operations;
accordingly the quality of the extracted part-level textual features is vulnerable to the change in visual domain.
By contrast, SSAN extracts the part-level textual features with the help of WAM, which is independent from the visual modality during the inference stage.
Therefore, SSAN is more robust than SCAN in these four transfer tasks.

Second, SSAN significantly outperforms all existing approaches in the ST setting in terms of all metrics.
Specifically, SSAN outperforms the most recent method MAN~\cite{jing2020cross} by 6.0\%, 6.0\%, 2.1\%, and 12.3\% in terms of Rank-1 accuracy on the four transfer tasks, respectively.
The superior performance of SSAN is not only due to the extraction of part-level features, but also due to its robustness to the change of visual domain during the inference stage.
These experimental results justify the robustness of SSAN for text-to-image ReID.

\subsection{Qualitative Results}

\begin{figure}
\begin{center}
\includegraphics[width=1.0\linewidth]{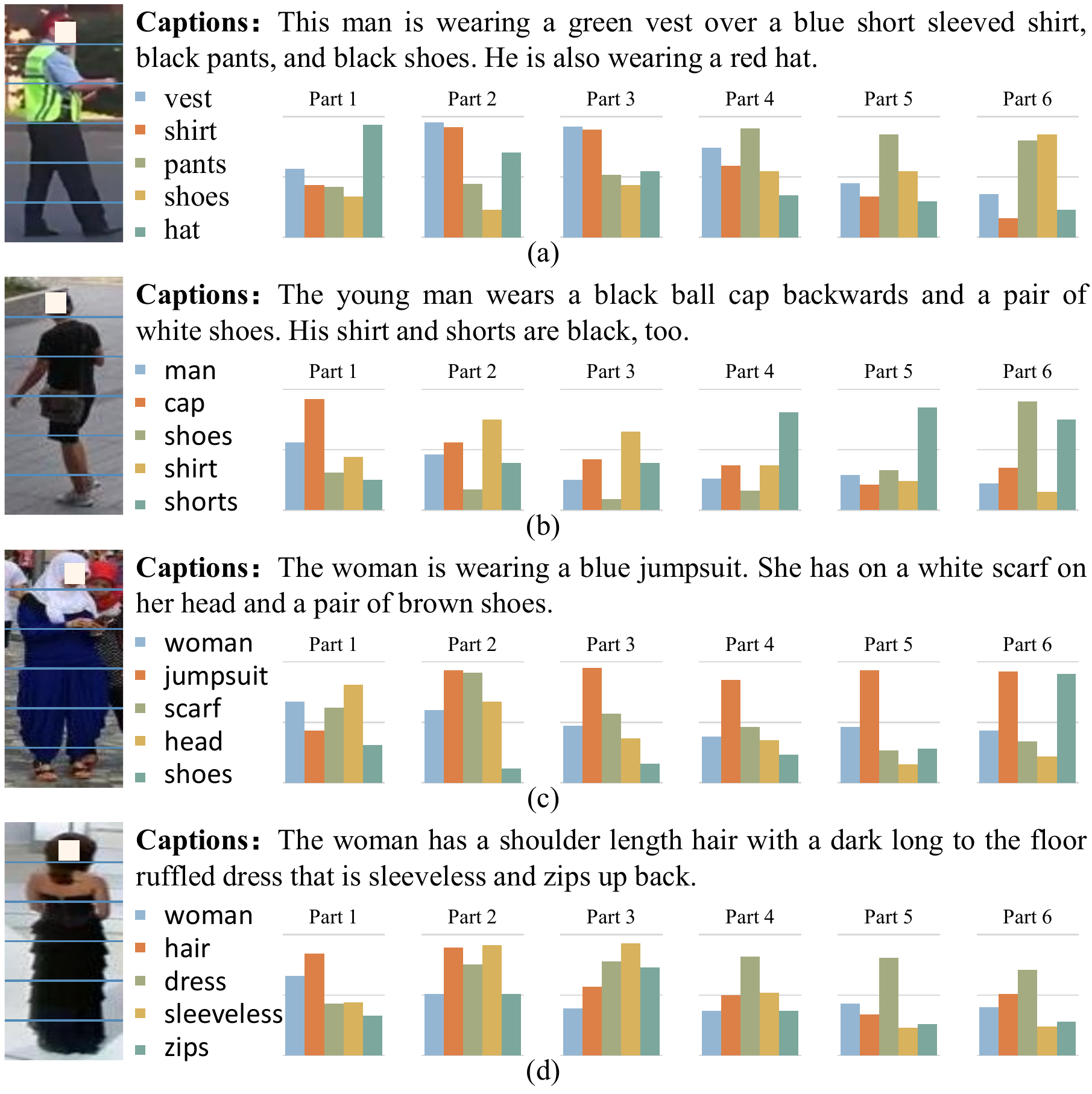}
\end{center}
   \caption{Illustration of WAM's prediction for one word that corresponds to fixed body parts.}
\label{fig:qua_grad_1}
\end{figure}

\begin{figure}
\begin{center}
\includegraphics[width=1.0\linewidth]{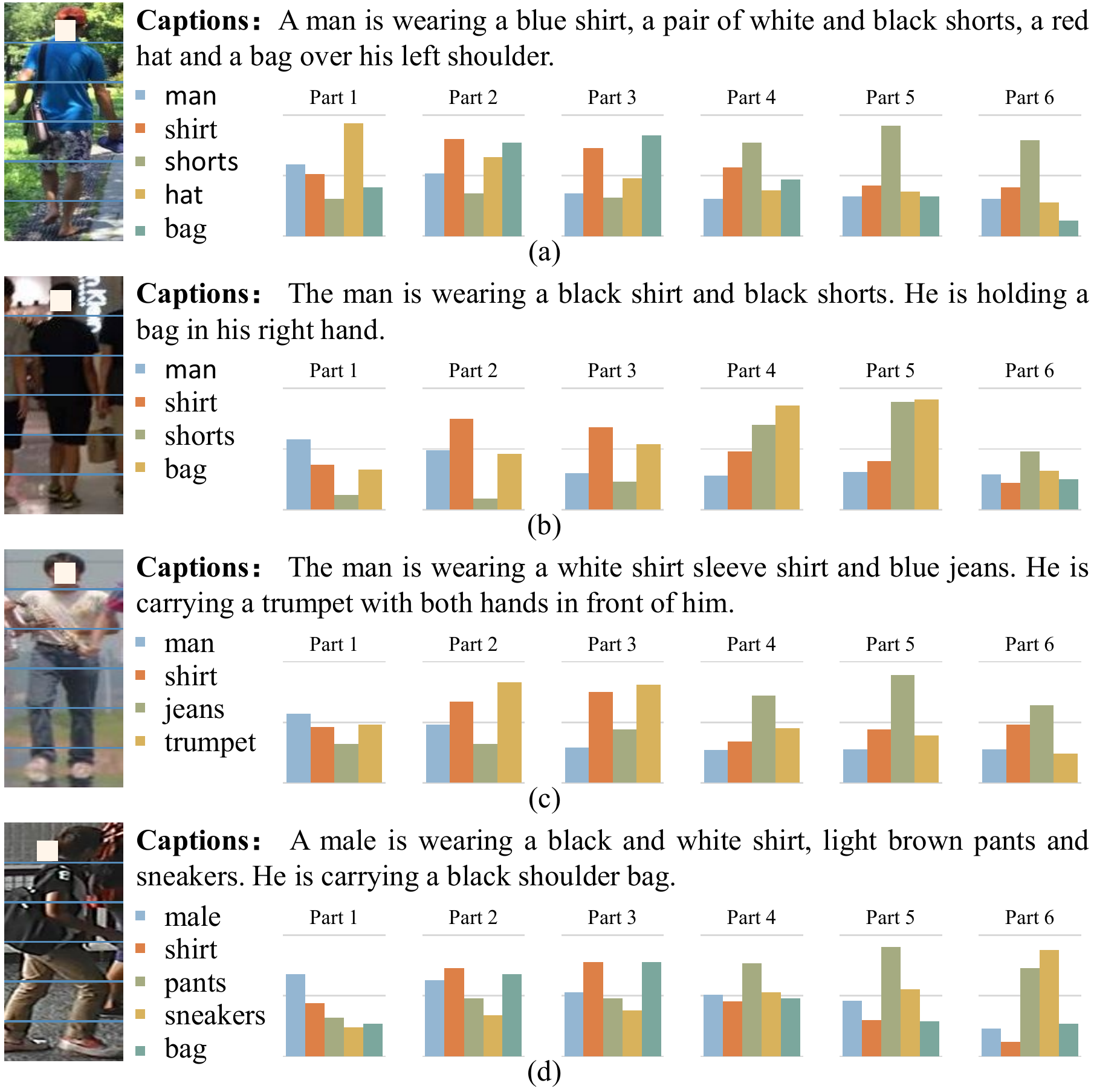}
\end{center}
   \caption{Illustration of WAM's prediction for one word that corresponds to object with flexible positions.}
\label{fig:qua_grad_2}
\end{figure}

As mentioned in Section~\ref{setion3.3}, WAM utilizes the relatively well-aligned visual body parts as supervision and the contextual cues in textual description to infer which part one word corresponds to. Therefore, WAM not only makes correct predictions for words that correspond to fixed body parts, but also makes reasonable predictions for words that correspond to objects with flexible positions. Here we provide the qualitative results about these two abilities in Fig.~\ref{fig:qua_grad_1} and Fig.~\ref{fig:qua_grad_2}, respectively. Each score denotes the probability that one word belongs to a certain body part. From the figure, we can make the following observations.

First, supervised by the relatively well-aligned body parts in images, WAM makes correct predictions for words that correspond to fixed body parts. For example, as illustrated in Fig.~\ref{fig:qua_grad_1}(a)-(b), the word ``shirt'' has high scores for the second and third parts. By contrast, both of the word ``jumpsuit''  in Fig.~\ref{fig:qua_grad_1}(c) and the word ``dress''  in Fig.~\ref{fig:qua_grad_1}(d) have high scores on the second to the fifth parts. The above qualitative results are in line with our common sense, which verify the reliability of WAM's predictive ability for the words that correspond to fixed body parts.

Second, by utilizing the contextual cues in textual description, WAM also makes reasonable predictions for words that correspond to objects with flexible positions. For example, the word ``bag'' in the phrase ``a bag over his left shoulder'' has high scores on the second and third parts in Fig.~\ref{fig:qua_grad_2}(a). By contrast, the word ``bag'' in the phrase ``a bag in his right hand'' has high scores on the fourth and fifth parts in Fig.~\ref{fig:qua_grad_2}(b). The above qualitative results demonstrate the effectiveness of WAM's predictive ability for the words that correspond to objects with flexible positions.

\section{Conclusion} \label{Conclusion}
In this paper, we propose a novel model, named SSAN, to automatically extract semantically aligned features from the visual and textual modalities for text-to-image ReID. Specifically, we introduce a word attention module that reliably attends to part-relevant words. This enables SSAN to automatically extract part-level features for the two modalities by shared $1\times1$ Conv layers. We further propose a multi-view non-local network to capture the relationships between body parts. Moreover, to overcome the large intra-class variance problem in textual descriptions, we propose a CR loss including both strong and weak supervision terms. Finally, to expedite research in text-to-image ReID, we build a new database that features identity-centric and fine-grained textual descriptions. Extensive experiments on two databases demonstrate the effectiveness of SSAN.

\ifCLASSOPTIONcaptionsoff
  \newpage
\fi



\bibliographystyle{IEEEtran}
\bibliography{references}
\end{document}